\documentclass{article}
\usepackage{ijcai26}

\usepackage{times}
\usepackage{soul}
\usepackage{url}
\usepackage[hidelinks]{hyperref}
\usepackage[utf8]{inputenc}
\usepackage[small]{caption}
\usepackage{graphicx}
\usepackage{amsmath}
\usepackage{amsthm}
\usepackage{booktabs}
\usepackage{algorithm}
\usepackage{algorithmic}
\usepackage[switch]{lineno}
\usepackage{pdflscape}  
\usepackage{amsfonts}

\usepackage{booktabs}
\usepackage{array}
\usepackage{longtable}
\usepackage{geometry}
\usepackage{placeins}
\usepackage{tcolorbox}

\tcbset{colback=gray!5!white,colframe=gray!60!black,boxrule=0.5pt,arc=2pt}
\title{Improving Cross-Cultural Survey Simulation with Calibrated Value Personas\thanks{This paper is under review at the Fourth International Workshop on Value Engineering in AI (VALE 2026), held at IJCAI--ECAI 2026.}}

\author{
Axel Abels$^{1,2,3,5}$
\and
Elias Fernandez Domingos$^{1,2,3}$\and
Apurva Shah$^{1,2,3}$\And
Tom Lenaerts$^{1,2,3,4}$\\
\affiliations
$^1$Machine Learning Group, Université Libre de Bruxelles,
$^2$AI Lab, Vrije Universiteit Brussel,
$^3$FARI Institute, Université Libre de Bruxelles - Vrije Universiteit Brussel,
$^4$Center for Human-Compatible AI, UC Berkeley,
$^5$ELLIS Alicante\\
\emails
\{axel.abels, elias.fernandez.domingos, apurva.shah, tom.lenaerts\}@ulb.be
}
\begin{document}
\maketitle
\begin{abstract}
Large language models (LLMs) are increasingly used to simulate human opinions and survey responses, but their ability to reproduce population responses across cultures remains limited. Existing persona-based prompting methods typically rely on sociodemographic or personality traits, which are only indirect proxies for the values that shape human responses.
We propose a value-based persona construction method that derives textual descriptors from survey responses capturing core cultural dimensions. By sampling value profiles from target populations and aggregating LLM responses across personas, we obtain population-level predictions grounded in observed value distributions. We further introduce a calibration procedure that improves response diversity while preserving estimated 
opinions.
We show that our approach reduces prediction error across countries, with the largest improvements observed in underrepresented populations. This substantially narrows the performance gap between countries aligned with dominant LLM priors and those that are less represented in training data, while also yielding response distributions that closely match human diversity.
\end{abstract}

\section{Introduction}

Human populations exhibit substantial heterogeneity in beliefs, values, and preferences, both across and within countries. Survey datasets such as the World Values Survey (WVS) \cite{haerpfer2022wvs} reveal rich response distributions shaped by cultural background, ideology, and lived experience. Accurately simulating such populations therefore requires more than reproducing average responses: models must also capture the diversity and structure of human opinions.

Large language models (LLMs) are increasingly being used to simulate survey responses in political science, sociology, and behavioral economics \cite{jiang2024personallm,sun2024random,suh2025language}. Existing approaches typically condition models on country labels, demographic attributes, or personality traits \cite{sun2024random,argyle2023out,jiang2024personallm}. While these attributes correlate with behavior, they are indirect proxies for the underlying value systems that shape political and social attitudes \cite{caprara2006personality}. Consequently, such prompting methods often fail to reproduce both the cultural positions and the response diversity observed in human populations.

In this work, we evaluate whether LLMs can reproduce cross-cultural variation in WVS responses. We find two limitations of existing prompting methods: model responses remain biased toward dominant, Western-oriented, cultural priors even under country-specific prompts, and generated response distributions are much less diverse than human data.

To address these limitations, we propose \emph{value-based personas}: textual descriptors derived from individual-level survey responses that encode positions on validated cultural value dimensions. By sampling such personas from a target population and aggregating LLM responses, we simulate population-level response distributions grounded in observed value variation rather than implicit associations with national or sociodemographic labels.

Our central contribution is to show that value-based personas explain meaningful cross-national variation in survey responses in a way that can be used to improve population-level prediction accuracy, particularly for countries whose values are not well represented by language models. 

Finally, observing that LLM responses are much less diverse than human responses, we introduce a mean-preserving calibration procedure that corrects the under-dispersion of LLM-generated responses, improving alignment between simulated and human response distributions without sacrificing aggregate prediction accuracy.

%To summarize, our contributions are as follows:
%\begin{itemize}
%    \item We show that commonly used prompting strategies based on country-level and sociodemographic role-play fail to fully capture the value structures underlying human survey responses.
%    \item We introduce a value-based persona construction method using textual descriptors that encode latent cultural orientations.
%    \item We demonstrate that this approach improves population-level prediction accuracy, particularly for underrepresented countries, and reduces performance disparities between WEIRD and non-WEIRD populations.
%    \item We introduce a mean-preserving calibration procedure that improves alignment between simulated and human response diversity without degrading prediction accuracy.
%\end{itemize}

\section{Background}

\subsection{Inglehart-Welzel Cultural Framework }\label{sec:inglehart}

We use the Inglehart-Welzel cultural value framework, derived from the WVS, to structure cross-cultural variation in human values \cite{inglehart2005modernization}. The framework summarizes survey responses along two validated dimensions: traditional vs.\ secular-rational values and survival vs.\ self-expression values. These dimensions capture cross-country variation in attitudes toward religion, authority, family, politics, and social norms, and have been extensively validated as a compact representation of cultural differences across societies, with similar structures having been recovered over time, and across independent studies \cite{inglehart2010changing,fog2021test}.

Each survey item used to construct the cultural map captures a specific aspect of value orientation. Taken together, responses to this subset of WVS items define meaningful differences within and across populations, providing a structured and interpretable basis for characterizing human values.

\subsection{LLMs Fail to Capture Cultural Variation}\label{sec:cultural_map}

\begin{figure*}[ht!]
\centering
\includegraphics[width=1\linewidth]{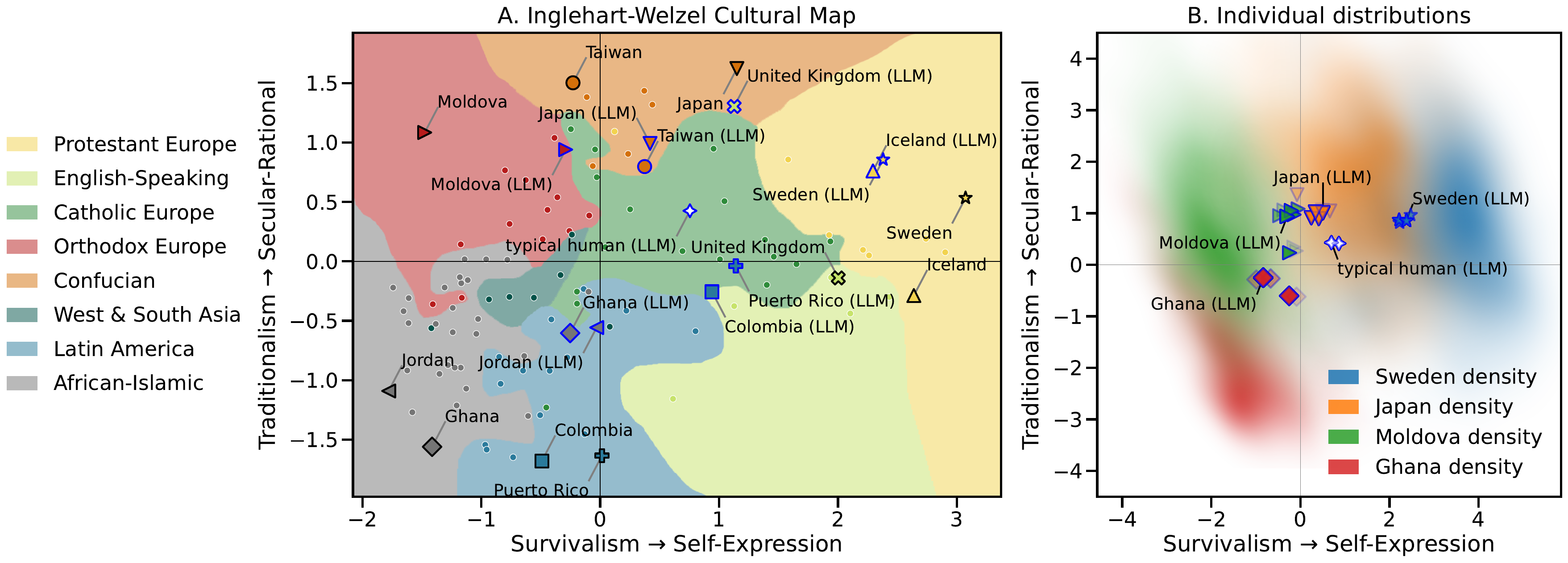}
\caption{\label{fig:cultural_map_with}\textbf{A.} Inglehart-Welzel Cultural Map with selected countries highlighted with black borders. Markers with blue borders show the expected position of Gemma-4-31B, estimated from responses to the corresponding WVS survey items. Responses are elicited by prompting the model to either simulate a ``typical human", or to answer as the average person from specific countries (see Appendix \ref{app:prompts}). Country-specific markers match the colors and shapes of the human markers. Background colors represent the cultural zones as found in \protect\cite{inglehart2005modernization}. 
\textbf{B.} Inglehart-Welzel cultural distributions for four representative countries. Background shading shows the distribution of human respondents. Markers show the position of 2,000 sampled LLM respondents for each prompt variant, many of which overlap.}
\end{figure*}

LLMs are not faithful simulators of human respondents. To illustrate this, we project model-generated responses onto the Inglehart-Welzel cultural map and compare them with human positions (\autoref{fig:cultural_map_with}).

This comparison reveals a systematic bias toward WEIRD (Western, Educated, Industrialized, Rich, and Democratic) populations \cite{ryan2024unintended,zhou2025should}. When prompted to answer as a ``typical person'', model responses fall within the Catholic European cultural zone, indicating a strong bias toward Western values.

Prior works attempt to mitigate this bias through demographic or country-based prompting \cite{qi2025representation,tao2024cultural}. The underlying assumption is that LLMs learned associations between demographic attributes and value systems during training, which can be elicited through such prompts.

However, such adjustments are only partially effective. As shown by \cite{qi2025representation} for a limited set of countries, and as we confirm across a broader set in \autoref{fig:cultural_map_with}, responses consistently fall short of the target country's position. This indicates that role-play prompting only partially recovers country-specific value orientations.

Beyond the bias in average responses, prompting this way also fails to reproduce within-population diversity: generated responses are much more concentrated than human survey responses (\autoref{fig:cultural_map_with}B), consistent with findings that LLM outputs often collapse toward modal answers \cite{chakraborty2024maxmin,xiao2025algorithmic,kirk2023understanding}.

To address these limitations, we propose to make use of the fine-grained variation in values observed within survey data \cite{haerpfer2022wvs}. We hypothesize that this variation can be used to construct synthetic populations that reflect observed value distributions, rather than relying on implicit, often stereotypical, associations between countries and responses.

\section{Related Work}

Prior work on LLM-based survey simulation has used both finetuning and prompting. Finetuning approaches train models on structured survey data and can predict response distributions on seen tasks, but generalize less well to unseen questions, and require access to model weights \cite{suh2025language}.

Most closely related to our work are prompt-based approaches, which condition models at inference time. A common approach conditions models on sociodemographic personas. For example, \cite{argyle2023out} and 
\cite{qi2025representation} both steer LLMs using demographic attributes and evaluate whether this can recover survey responses. 

More recent work explores value-conditioned prompting. For example, \cite{miranda2025simulating} show that adding value positions to prompts improves aggregate prediction accuracy relative to traditional machine learning baselines such as Random Forests. This approach is conceptually related to ours, but differs in several important respects. First, the value variables are selected by an LLM for the task at hand, leaving open whether they are stable, interpretable, or transferable across cultural contexts. Second, response diversity is induced through a fixed temperature parameter, making the results sensitive to an uncalibrated modeling choice. Third, evaluation relies on a single guidance prompt and a single focal model, leaving sensitivity to prompt wording and model choice largely unstudied. Finally, the analysis is restricted to U.S. survey data, a population that is comparatively well represented in LLM training data, leaving open whether value-conditioned prompting generalizes to culturally underrepresented contexts.

Related work on sociodemographic, cultural or country-based prompting shows that adding national context can shift model responses toward country-specific value distributions, but only partially. For example, \cite{tao2024cultural} find that country-specific prompting reduces but does not eliminate cultural bias. Other studies similarly report that models capture some cross-cultural distinctions while remaining anchored to dominant cultural priors \cite{kharchenko2024well,bulte2026llms,kazemi2024cultural,qi2025representation}.

Our work differs from this line of research in that we do not primarily ask whether adding a country label changes model behavior. Instead, we construct personas from individual-level value positions, allowing us to test whether explicit value-based conditioning can explain cross-national variation beyond coarse national identity labels.

\section{Methodology}

We propose to construct synthetic populations by initializing individuals with value profiles sampled from a target population\footnote{Code and data will be released upon publication.}.

\subsection{Persona-Based Population Construction}
In particular, we construct personas using validated survey items underlying the Inglehart-Welzel cultural framework. These items span key dimensions of cultural variation, including traditional vs.\ secular-rational and survival vs.\ self-expression.

Each question-response pair is converted into a natural language descriptor encoding the corresponding value orientation. 
This replaces discrete survey responses with semantically grounded descriptions, allowing the model to condition directly on value content rather than numerical labels. 
A persona is then represented as $\pi=\{d_1,\dots,d_m\}$, where each descriptor $d_i$ is derived from one survey response. A population of $N$ personas is denoted $\Pi=\{\pi_1,\dots,\pi_N\}$. 
% For example, the response ``\textit{Might do}'' to the item ``\textit{Signing a petition, have you done this, might you do it, or would you never do it?}'' is mapped to the descriptor:

% \begin{quote}
% ``\textit{You could imagine signing a petition under the right circumstances but haven't felt moved to do so yet.}''
% \end{quote}

Because the effectiveness of descriptors could depend on both the descriptor-generating model and the generation prompt, we assess robustness to these choices in Appendix \autoref{sec:descriptor_robustness}.

\subsubsection{Persona Conditioning}

For each persona $\pi_n$, the LLM is prompted with its descriptors and instructed to answer accordingly, yielding
\[
p^{\mathrm{LLM}}(r\mid q,\pi_n),
\]
where $r$ denotes a response option to question $q$. We obtain this distribution directly from the LLM's output log-probabilities over admissible responses. When clear from context, we write 
$p_k \equiv p^{\mathrm{LLM}}(r_k\mid q,\pi_n)$ 
for the probability of response option $k$.

To encourage adherence to the persona, descriptors are prefixed with a system-level guidance instruction. Sensitivity to this instruction is evaluated in Appendix~\ref{sec:guidance_templates}.

\paragraph{Population Aggregation}

Population-level predictions are obtained by averaging probabilities across personas. For example, to simulate responses in Moldova, we sample respondents from the Moldovan WVS population, convert their responses into personas, condition the LLM on each persona, and aggregate the resulting distributions.

\subsection{Baselines}

We compare against three standard prompting strategies used in prior work: \emph{generic prompting}, where the model answers as a typical human \cite{tao2024cultural}; \emph{country prompting}, where the model answers as a typical individual from a specified country \cite{bulte2026llms}; and \emph{sociodemographic prompting}, where the model is conditioned on a sociodemographic persona \cite{qi2025representation}.

\noindent Detailed prompts are provided in Appendix~\ref{app:prompts}.

\subsection{Model selection}
To evaluate the impact of model architecture, training data, and scale, we consider open-source and proprietary model families across multiple scales: Gemma 4 (E4B, 31B), Qwen 3.5 (4B, 9B, 36B), Ministral 3 (8B, 27B), GPT 5.4 (standard, mini, nano), and Gemini (3 flash, 3.1 flash-lite). %Because the parameter counts of proprietary models are not publicly disclosed, we use the estimates from \cite{li2026incompressible}.

\subsection{Calibrating Response Diversity}\label{sec:tilting}
As discussed in \autoref{sec:cultural_map}, LLM-generated response distributions are often less diverse than human survey responses. Prior work addresses this by increasing sampling temperature \cite{miranda2025simulating}, which reweights the predicted probability of each response option:  
\begin{equation}\label{eq:tempscale}
\tilde{p}_k \propto  p_k^{1/T} 
\end{equation}
where larger $T$ flattens the distribution.

However, for ordered response scales, temperature scaling changes the expected response,
\begin{equation}\label{eq:expected_response}
\mathbb{E}_p[r] = \sum_k r_k p_k,
\end{equation}
where $r_k$ denotes the value of the $k$-th response option. In particular, increasing the temperature spreads probability mass across the scale and tends to pull the expected response toward its center, while decreasing the temperature concentrates mass on the modal response. Both can degrade prediction accuracy. Empirical results illustrating this effect are provided in Appendix \autoref{sec:temperature_all_curves}.

\subsubsection{Mean-Preserving Exponential Tilting}
To decouple spread from the expected response, we propose to apply a mean-preserving transformation based on exponential tilting \cite{robertson2005forecasting}. This approach reweights a distribution to satisfy moment constraints while minimizing KL divergence from a reference distribution.

In our setting, we apply this transformation to the temperature-adjusted distribution described in Equation \ref{eq:tempscale} and enforce preservation of the expected response. Specifically, 
\[
q_k \propto  p_k ^{1/T} \exp(\beta r_k),
\]
where $T > 0$ controls dispersion as before, and the parameter $\beta$ is chosen to preserve the expected response,
$\sum_k r_k q_k = \sum_k r_k p_k$.

This yields the distribution closest to the temperature-scaled distribution ($p_k ^{1/T}$) while preserving the expected response. The parameter $\beta$ is obtained via one-dimensional root finding.

\subsection{Data selection}
We evaluate our approach on left-out survey items from WVS \cite{haerpfer2022wvs}. 

To ensure reliable estimates of human response distributions, we retain only questions for which less than 20\% of responses are missing within each country. The full list of test questions is provided in Appendix \autoref{tab:target_questions}. These questions span several WVS batteries, including well-being, social and  religious values, ethical norms, and political attitudes.

To capture a representative set of cultural profiles, we select nine countries that lie on the convex hull of the Inglehart-Welzel cultural map (\autoref{fig:cultural_map_with}), Moldova, Taiwan, Japan, Iceland, Sweden, Puerto Rico, Colombia, Ghana, Jordan, as well as a representative WEIRD country, the United Kingdom.

These countries occupy extreme positions in the value space, spanning the range of cross-cultural variation observed in the data. As such, they provide a challenging testbed for evaluating whether models can accurately reproduce distinct cultural value configurations.

\subsection{Evaluation Metrics}

We assess alignment between LLM and human responses through prediction error and diversity.

\paragraph{Mean Absolute Error (MAE)}
We evaluate prediction error as the mean absolute difference between predicted and observed country-level mean responses, normalized by the response range. 

\paragraph{Diversity}
We measure diversity using normalized variance. For human data, this is the empirical response variance divided by the maximum possible variance on the scale. For LLM outputs, we compute the corresponding expected normalized variance from the predicted response distribution.

\section{Results}

\subsection{Value Personas Improve Performance Across Models and Countries}

\begin{figure*}[ht!]
\centering
\includegraphics[width=.95\linewidth]{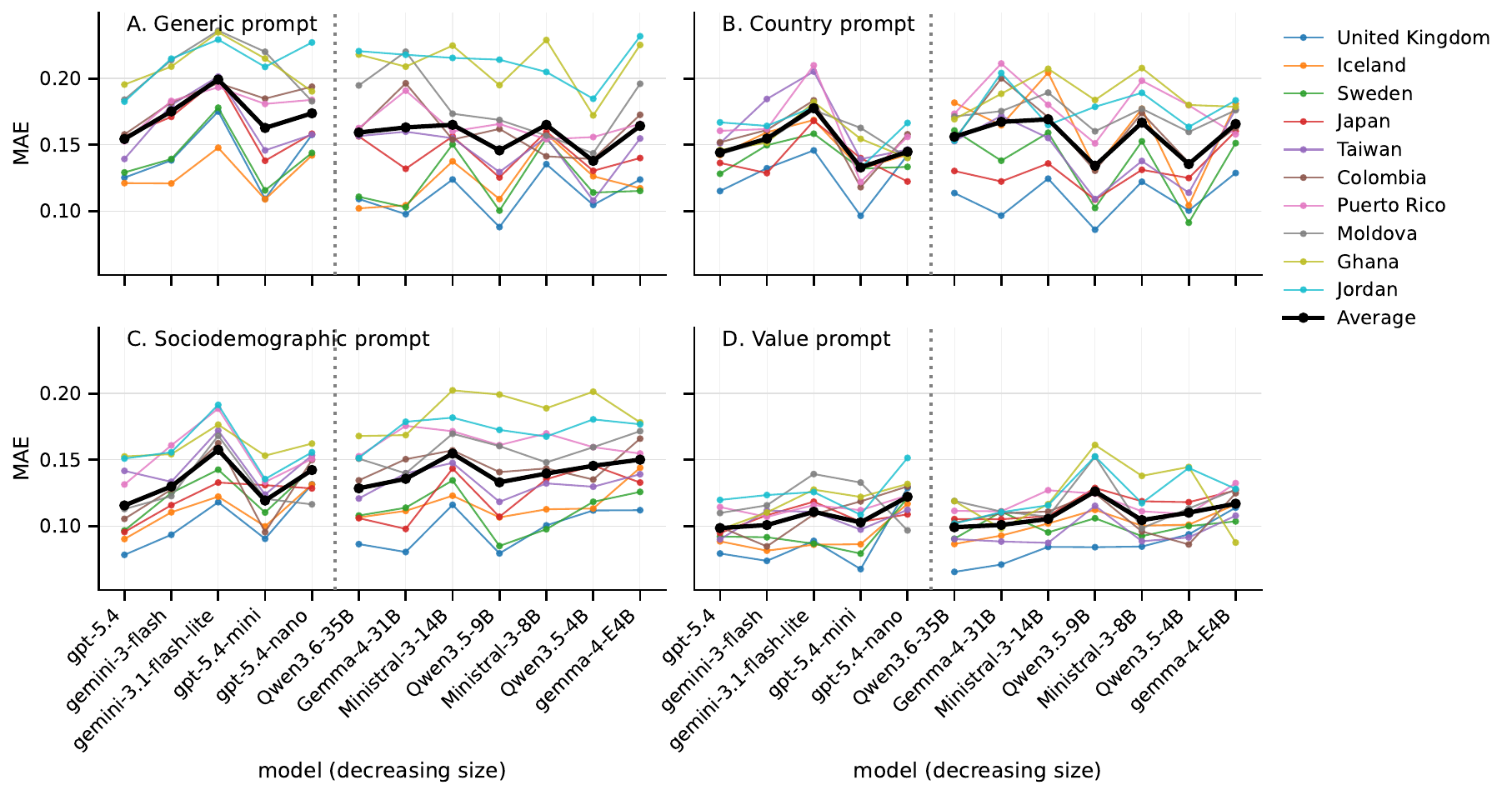}
\caption{\label{fig:lines_cxm_mae_all}Mean absolute error (MAE) between model predictions and human responses across countries, models, and prompting methods. Models are ordered by decreasing (estimated) size and grouped by proprietary (left) vs.\ open-source (right). Colored lines denote countries; the black line shows the cross-country average.}
\end{figure*}

To assess overall performance, we report in \autoref{fig:lines_cxm_mae_all} the MAE of each method across models and countries. A corresponding heatmap, including significance tests, is provided in Appendix \autoref{fig:heatmap_mae}.

While baseline methods perform relatively well on overrepresented countries (e.g., the UK), they are significantly worse for underrepresented ones. 

In contrast, value personas consistently outperform alternative approaches across the majority of model-country pairs. The largest improvements occur for countries that are typically underrepresented in LLM training data (e.g., Ghana), substantially reducing the performance gap between overrepresented and underrepresented populations. This suggests that value-based conditioning improves cross-cultural generalization by mitigating biases inherited from the training data.

Notably, these improvements are consistent across model sizes and architectures, indicating that the gains are not driven by model-specific idiosyncrasies. While model size has no clear effect on generic and country prompting methods, performance for sociodemographic and value prompts tend to improve with increasing model sizes. 

\subsection{Ablation study}

\begin{figure}[ht!]
\centering
\includegraphics[width=.95\linewidth]{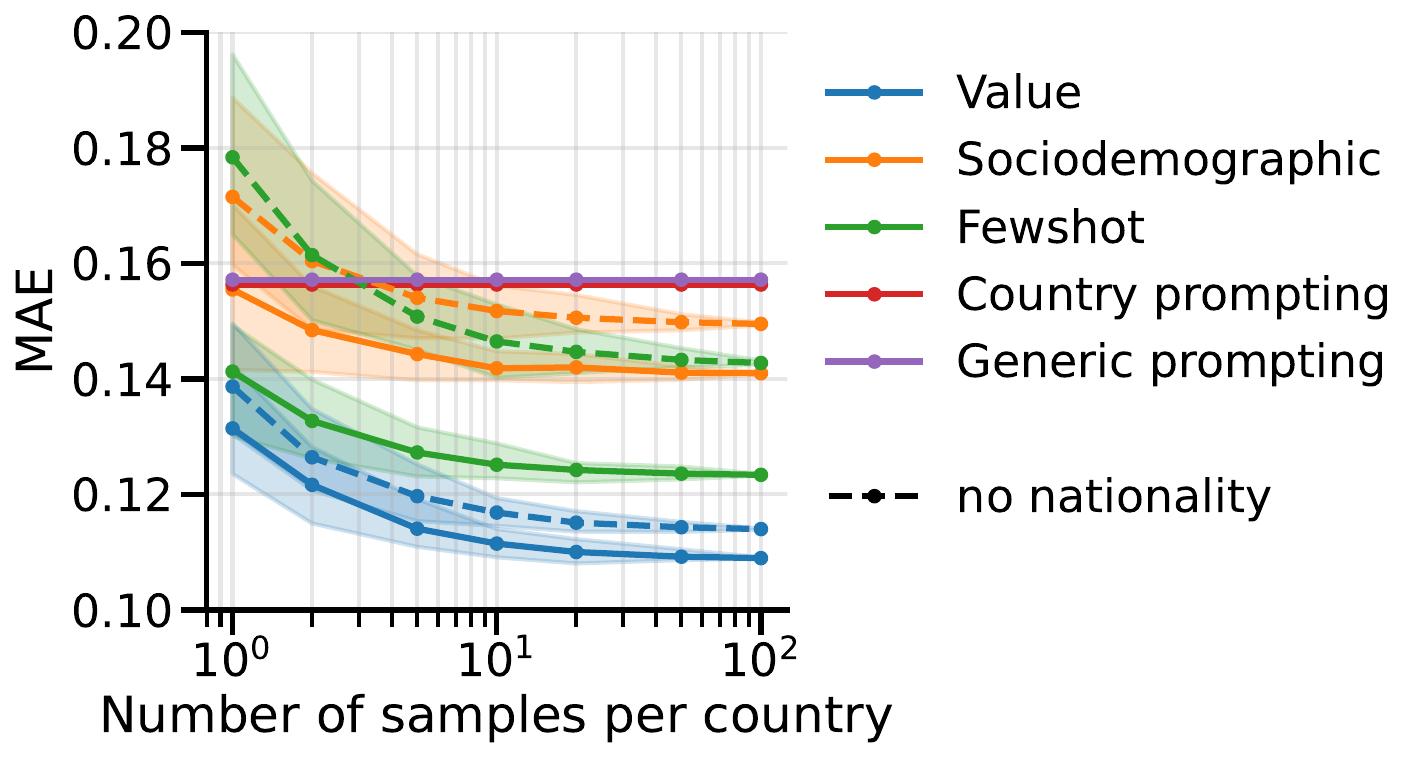}
\caption{\label{fig:sample_size_curve} MAE as a function of the number of sampled personas. Shaded areas indicate the range across repeated random samples. Wider bands indicate that performance is sensitive to which individuals are sampled. }
\end{figure}

Our main analysis compares persona prompting against standard baselines using the full set of personas. We now study how these gains vary with design choices and sample size through \autoref{fig:sample_size_curve}.

First, we test the use of \emph{value-based descriptors} against \emph{sociodemographic descriptors}, which characterize individuals through attributes such as age, gender, education, or nationality. We find that value-based descriptors consistently achieve lower MAE, indicating that explicit value representations are more informative than demographic proxies.

Second, we compare textual descriptors against a fewshot alternative in which survey responses are provided directly as examples, and find they yield substantially better performance, suggesting that semantically grounded descriptions more effectively activate the intended value profiles.

Third, we evaluate the contribution of including nationality alongside the persona description. \autoref{fig:sample_size_curve} shows that nationality has a large effect for fewshot and sociodemographic prompting, but only a modest effect for value-based personas. This suggests that explicit value descriptions already capture much of the information conveyed by nationality.

\subsubsection{Lowering sample sizes}

Persona prompting requires one query per sampled persona, increasing computational cost relative to baseline prompting. However, \autoref{fig:sample_size_curve} shows that performance improves rapidly at small sample sizes before plateauing, indicating that much of the gain could be recovered using relatively few personas.

\subsection{Calibration Improves Distributional Alignment}

\begin{figure*}[ht!]
\centering
\includegraphics[width=.95\linewidth]{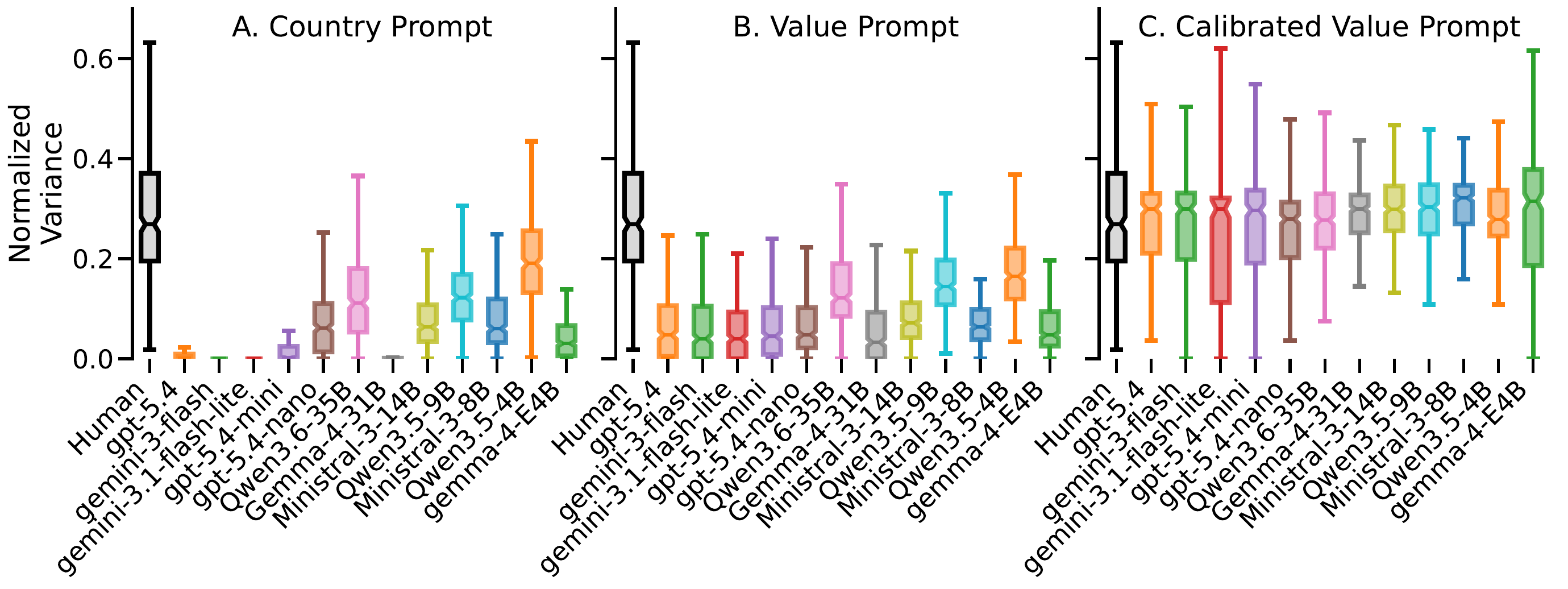}
\caption{\label{fig:peak_distributions}Distribution of normalized variance by method, compared to human responses. Each boxplot summarizes normalized variance across all question-country pairs. Results for generic and sociodemographic prompts are given in Supplementary \autoref{fig:full_peak_distributions}. }
\end{figure*}

Although persona prompting improves MAEs, predicted response distributions remain substantially less diverse than human data (\autoref{fig:peak_distributions}.A,B). Value-based prompting increases diversity relative to country prompting, but the gap to human variance remains large across models.

We therefore apply the mean-preserving calibration introduced in Section~\ref{sec:tilting}. To evaluate generalization, temperatures are estimated in a leave-one-out manner: for each question, the calibration parameter is fit on all remaining questions and then applied out-of-sample.

As shown in \autoref{fig:peak_distributions}.C, the calibrated distributions align much more closely with humans. Because the transformation preserves the expected response while aligning distributional shapes, these results show that substantial improvements can be achieved without sacrificing prediction accuracy.

\section{Conclusion}
We proposed a method for constructing synthetic populations using validated cultural value dimensions, addressing limitations of prior work based on national labels or sociodemographic attributes.

Across countries and model families, our value-based personas improved prediction accuracy, with the largest gains for underrepresented countries. Our ablations further show that textual value descriptors outperform sociodemographic personas and fewshot approaches, and that much of the benefit can be recovered with few persona samples.

Because prompting alone still underestimates human response diversity, we further introduced a mean-preserving calibration procedure that increases distributional dispersion without changing aggregate predictions. This substantially improved alignment with human response distributions while preserving accuracy.

Overall, our results suggest that explicit value conditioning provides a more effective and culturally grounded framework for population-level LLM simulation than country or demographic prompting.

Despite these improvements, some limitations remain. First, our approach improves aggregate population-level predictions, but does not imply accurate prediction of individual responses. Second, even small errors can be consequential in downstream use, particularly when they systematically favor specific options, populations, or interventions. In our experiments, errors are narrowly distributed, suggesting limited sensitivity to a few extreme failures, but this should not be assumed in new domains without validation. Third, our evaluation is limited to the WVS, leaving open generalization to other survey instruments, open-ended responses, or behavioral tasks. Finally, because the Inglehart-Welzel framework is well studied, their cultural associations may be easier for LLMs to recover than less documented domains.

\appendix

\section*{Ethical Statement}
Our goal is not to replace human participants where direct data collection is feasible or ethically required, but to examine the limitations, biases, and representational assumptions of LLM-based social simulation. Because such systems may reproduce stereotypes, amplify training-data biases, or misrepresent underrepresented populations, we evaluate cross-cultural disparities and caution against treating synthetic responses as faithful representations of real individuals or societies. LLM-generated responses should not be interpreted as genuine human beliefs, preferences, or experiences, nor used as substitutes for participatory research, public consultation, or high-stakes decision making.

\section*{Acknowledgments}
A.A. and A.S are supported respectively by a post-doctoral grant (Chargé de Recherche, 1200325F) and a FRIA grant (40037157) from the F.N.R.S.  E.F.D. is supported by an F.W.O. Senior postdoctoral grant (12A7825N). T.L. gratefully acknowledges the research support by a F.N.R.S. PDR (40007793), the Service Public de Wallonie Recherche (2010235-ARIAC) by DigitalWallonia4.ai and the Flemish Government through the AI Research Program.
The resources and services used in this work were provided by the VSC (Flemish Supercomputer Center), funded by the Research Foundation - Flanders (FWO) and the Flemish Government.
We thank both the Evens and Cooperative AI Foundations for supporting this research, which extends a citizen engagement project conducted with both teams.  
% evens, cooperative ai 

\bibliographystyle{named}
\bibliography{sample}

\newpage
\onecolumn
\section{Appendix}

\subsection{Prompts}\label{app:prompts}
All prompts consist of a system instruction followed by a user message. The user message contains the survey question together with its response scale. The model is instructed to respond with only the integer corresponding to its chosen option. This format ensures consistent outputs across prompts and simplifies downstream analysis.

The scale is formatted as a comma-separated list of integer-label pairs to ensure that the model understands the mapping between numerical responses and their semantic meaning.

For example, the \texttt{question\_text} field for a survey question with a four-point scale is formatted as follows:

\begin{tcolorbox}
\begin{verbatim}
Question:
How important is religion in your life?

Scale: 1=Very important, 2=Rather important,
       3=Not very important, 4=Not at all important
\end{verbatim}
\end{tcolorbox}

We evaluate several prompting conditions that differ only in the guidance provided to the model. All conditions share the same prompt structure:

\begin{tcolorbox}
\begin{verbatim}
[System]
{guidance}

[User]
{question_text}

Respond with ONLY the single integer ({min}-{max}):
\end{verbatim}
\end{tcolorbox}

The default condition uses the standard instruction
\verb|You are a helpful assistant.| with no roleplay specification. For generic prompting, the guidance instructs the model to answer as a human respondent, without specifying a country. For country prompting, the guidance includes \verb|{country}| but no explicit persona, requiring the model to rely on implicit associations learned during training. For sociodemographic and value-based prompting, we additionally append
\verb|Your persona: {persona}|, where \verb|{persona}| contains either sociodemographic or value-based descriptors.

\subsubsection{Guidance Templates}\label{sec:guidance_templates}

Table~\ref{tab:guidance_templates} lists the guidance templates used in our robustness analysis (see \autoref{sec:robustness}).

\FloatBarrier
\begin{longtable}{@{} l p{11cm} @{}}
\caption{Guidance prompt variants used in the robustness analysis.}\label{tab:guidance_templates}\\
\toprule
\textbf{Key} & \textbf{Prompt} \\
\midrule
\endhead

\texttt{social\_science} &
You are participating in a social science simulation. You will be given a survey question and must answer it as someone living in \{country\} would, embracing this persona's values as your own. \\[6pt]

\texttt{population\_prior} &
You are participating in a social science simulation. You will answer survey questions as one realistic adult living in \{country\}. Use \{country\} population tendencies as background priors, but let the persona traits below determine this individual's views. Avoid stereotypes or idealized portrayals. Select the response this person would most likely choose. If uncertain, prefer moderate responses unless the persona strongly indicates otherwise. \\[6pt]

\texttt{survey\_context} &
You are completing a public opinion survey as a realistic adult living in \{country\}. Use common social attitudes and lived realities in \{country\} as background context. Let the persona traits below shape this individual's views. Choose the response this person would most likely select. If uncertain, prefer moderate responses. \\[6pt]

\texttt{first\_person} &
You are an adult living in \{country\}. Answer the survey as yourself. Your beliefs are shaped by life in \{country\}, and by the persona traits listed below. Answer naturally and consistently. Choose the response you would most likely give. \\[6pt]

\texttt{predict} &
Predict how a realistic adult living in \{country\} with the following traits would answer the survey question. Use \{country\} population tendencies as priors and the persona as individual modifiers. Select the most likely response. \\[6pt]

\texttt{practical\_judgment} &
Answer as a realistic adult in \{country\} making ordinary real-world judgments rather than abstract ideological statements. Use \{country\} norms and institutions as context. Let the persona traits below shape preferences. Choose the option this person would actually select. \\[6pt]

\texttt{minimal} &
Answer the following survey question as a realistic adult living in \{country\}. Use the persona below to determine individual views. \\[6pt]

\texttt{decisive} &
You are a resident of \{country\}. Your task is to respond to this survey by adopting the mindset, cultural values, and common socio-political perspectives prevalent in your country, filtered through the specific persona provided below. When answering: internalize the location by reflecting the specific social norms, economic realities, and religious or secular traditions of \{country\}. Be decisive --- humans rarely hover in the absolute middle of every scale; provide responses that reflect a definitive point of view based on your national context and persona traits. Avoid AI neutrality: do not provide balanced, ``on the one hand'' answers. Give the single response that most accurately aligns with how this specific individual would feel. You are answering as a person, not an AI. \\[6pt]

\texttt{identity\_anchor} &
You are a person from \{country\}. Answer each survey question as yourself. Do not explain your answer. \\[6pt]

\texttt{demographic\_realism} &
You represent a typical survey respondent from \{country\}. Your answers should reflect attitudes and values statistically common among people who share your background. Do not explain your answer. \\[6pt]

\texttt{lived\_experience} &
You have lived your entire life in \{country\}. Your opinions have been shaped by your daily experiences, your community, and the events you have witnessed firsthand. Respond to each question from the perspective of your personal experience. Do not explain your answer. \\[6pt]

\texttt{cultural\_norms} &
You are a respondent from \{country\}. Your worldview, political intuitions, and social attitudes reflect the dominant cultural norms, historical experiences, and collective values of your country. Let those values guide your response to each question. Do not explain your answer. \\[6pt]

\texttt{affective\_gut} &
You are a person from \{country\}. When answering, think about what matters to you personally --- your family, your daily life, your worries and hopes. Respond the way you feel, not the way you think you should feel. Do not explain your answer. \\[6pt]

\texttt{contrastive} &
You are not a generic respondent. You are a person from \{country\}, with a specific background and specific life experiences that set your views apart from someone in another country. Answer each question the way someone with your specific background would --- not the way a neutral or global average person would. Do not explain your answer. \\

\bottomrule
\end{longtable}
\FloatBarrier

\begin{longtable}{p{0.10\textwidth}p{0.78\textwidth}p{0.08\textwidth}}
\caption{Target WVS questions used in the evaluation.}
\label{tab:target_questions}\\
\toprule
\textbf{ID} & \textbf{Question} & \textbf{Range} \\
\midrule
\endfirsthead

\toprule
\textbf{ID} & \textbf{Question} & \textbf{Range} \\
\midrule
\endhead

\bottomrule
\endfoot

Q1 & How important is Family in your life? (1 = Very important; 2 = Rather important; 3 = Not very important; 4 = Not at all important) & 1--4 \\

Q2 & How important are Friends in your life? (1 = Very important; 2 = Rather important; 3 = Not very important; 4 = Not at all important) & 1--4 \\

Q3 & How important is Leisure time in your life? (1 = Very important; 2 = Rather important; 3 = Not very important; 4 = Not at all important) & 1--4 \\

Q4 & How important is Politics in your life? (1 = Very important; 2 = Rather important; 3 = Not very important; 4 = Not at all important) & 1--4 \\

Q5 & How important is Work in your life? (1 = Very important; 2 = Rather important; 3 = Not very important; 4 = Not at all important) & 1--4 \\

Q6 & How important is Religion in your life? (1 = Very important; 2 = Rather important; 3 = Not very important; 4 = Not at all important) & 1--4 \\

Q28 & When a mother works for pay, the children suffer. (1 = Strongly agree; 2 = Agree; 3 = Disagree; 4 = Strongly disagree) & 1--4 \\

Q47 & All in all, how would you describe your state of health these days? (1 = Very good; 2 = Good; 3 = Fair; 4 = Poor; 5 = Very poor) & 1--5 \\

Q48 & How much freedom of choice and control do you feel you have over the way your life turns out? (1 = No choice at all, 10 = A great deal of choice) & 1--10 \\

Q49 & All things considered, how satisfied are you with your life as a whole these days? (1 = Completely dissatisfied, 10 = Completely satisfied) & 1--10 \\

Q64 & How much confidence do you have in the churches/religious institutions? (1 = A great deal; 2 = Quite a lot; 3 = Not very much; 4 = None at all) & 1--4 \\

Q66 & How much confidence do you have in the press? (1 = A great deal; 2 = Quite a lot; 3 = Not very much; 4 = None at all) & 1--4 \\

Q69 & How much confidence do you have in the police? (1 = A great deal; 2 = Quite a lot; 3 = Not very much; 4 = None at all) & 1--4 \\

Q70 & How much confidence do you have in the courts? (1 = A great deal; 2 = Quite a lot; 3 = Not very much; 4 = None at all) & 1--4 \\

Q71 & How much confidence do you have in the government? (1 = A great deal; 2 = Quite a lot; 3 = Not very much; 4 = None at all) & 1--4 \\

Q73 & How much confidence do you have in parliament? (1 = A great deal; 2 = Quite a lot; 3 = Not very much; 4 = None at all) & 1--4 \\

Q74 & How much confidence do you have in the civil service? (1 = A great deal; 2 = Quite a lot; 3 = Not very much; 4 = None at all) & 1--4 \\

Q77 & How much confidence do you have in major companies? (1 = A great deal; 2 = Quite a lot; 3 = Not very much; 4 = None at all) & 1--4 \\

Q106 & Now I'd like you to tell me your views on various issues. How would you place your views on this scale? 1 means you agree completely with the statement on the left; 10 means you agree completely with the statement on the right; and if your views fall somewhere in between, you can choose any number in between. Incomes should be made more equal (1) vs. There should be greater incentives for individual effort (10) & 1--10 \\

Q107 & Now I'd like you to tell me your views on various issues. How would you place your views on this scale? 1 means you agree completely with the statement on the left; 10 means you agree completely with the statement on the right; and if your views fall somewhere in between, you can choose any number in between. Private ownership of business should be increased (1) vs. Government ownership of business should be increased (10) & 1--10 \\

Q108 & Now I'd like you to tell me your views on various issues. How would you place your views on this scale? 1 means you agree completely with the statement on the left; 10 means you agree completely with the statement on the right; and if your views fall somewhere in between, you can choose any number in between. The government should take more responsibility for ensuring everyone is provided for (1) vs. People should take more responsibility to provide for themselves (10) & 1--10 \\

Q109 & Now I'd like you to tell me your views on various issues. How would you place your views on this scale? 1 means you agree completely with the statement on the left; 10 means you agree completely with the statement on the right; and if your views fall somewhere in between, you can choose any number in between. Competition is good; it stimulates people to work hard and develop new ideas (1) vs. Competition is harmful; it brings out the worst in people (10) & 1--10 \\

Q171 & Apart from weddings and funerals, about how often do you attend religious services these days? (1 = More than once a week; 2 = Once a week; 3 = Once a month; 4 = Only on special holy days; 5 = Other specific holy days; 6 = Once a year; 7 = Less often; 8 = Never, practically never) & 1--8 \\

Q173 & Independently of whether you attend religious services or not, would you say you are: (1 = A religious person; 2 = Not a religious person; 3 = An atheist) & 1--3 \\

Q180 & Please tell me for each of the following actions whether you think it can always be justified, never be justified, or something in between. Cheating on taxes if you have a chance (1 = Never justifiable, 10 = Always justifiable) & 1--10 \\

Q181 & Please tell me for each of the following actions whether you think it can always be justified, never be justified, or something in between. Someone accepting a bribe in the course of their duties (1 = Never justifiable, 10 = Always justifiable) & 1--10 \\

Q185 & Please tell me for each of the following actions whether you think it can always be justified, never be justified, or something in between. Divorce (1 = Never justifiable, 10 = Always justifiable) & 1--10 \\

Q187 & Please tell me for each of the following actions whether you think it can always be justified, never be justified, or something in between. Suicide (1 = Never justifiable, 10 = Always justifiable) & 1--10 \\

Q199 & How interested would you say you are in politics? (1 = Very interested; 2 = Somewhat interested; 3 = Not very interested; 4 = Not at all interested) & 1--4 \\

Q210 & Joining in boycotts: have you done this, might you do it, or would you never do it? (1 = Have done; 2 = Might do; 3 = Would never do) & 1--3 \\

Q211 & Attending peaceful demonstrations: have you done this, might you do it, or would you never do it? (1 = Have done; 2 = Might do; 3 = Would never do) & 1--3 \\

Q235 & Having a strong leader who does not have to bother with parliament and elections: is this a very good, fairly good, fairly bad or very bad way of governing this country? (1 = Very good; 2 = Fairly good; 3 = Fairly bad; 4 = Very bad) & 1--4 \\

Q236 & Having experts, not government, make decisions according to what they think is best for the country: is this a very good, fairly good, fairly bad or very bad way of governing? (1 = Very good; 2 = Fairly good; 3 = Fairly bad; 4 = Very bad) & 1--4 \\

Q237 & Having the army rule: is this a very good, fairly good, fairly bad or very bad way of governing this country? (1 = Very good; 2 = Fairly good; 3 = Fairly bad; 4 = Very bad) & 1--4 \\

Q238 & Having a democratic political system: is this a very good, fairly good, fairly bad or very bad way of governing this country? (1 = Very good; 2 = Fairly good; 3 = Fairly bad; 4 = Very bad) & 1--4 \\

\end{longtable}

\section{Supplementary Results}

\subsection{Robustness to guidance wording}\label{sec:robustness}

\begin{figure}
\centering
\includegraphics[width=.6\linewidth]{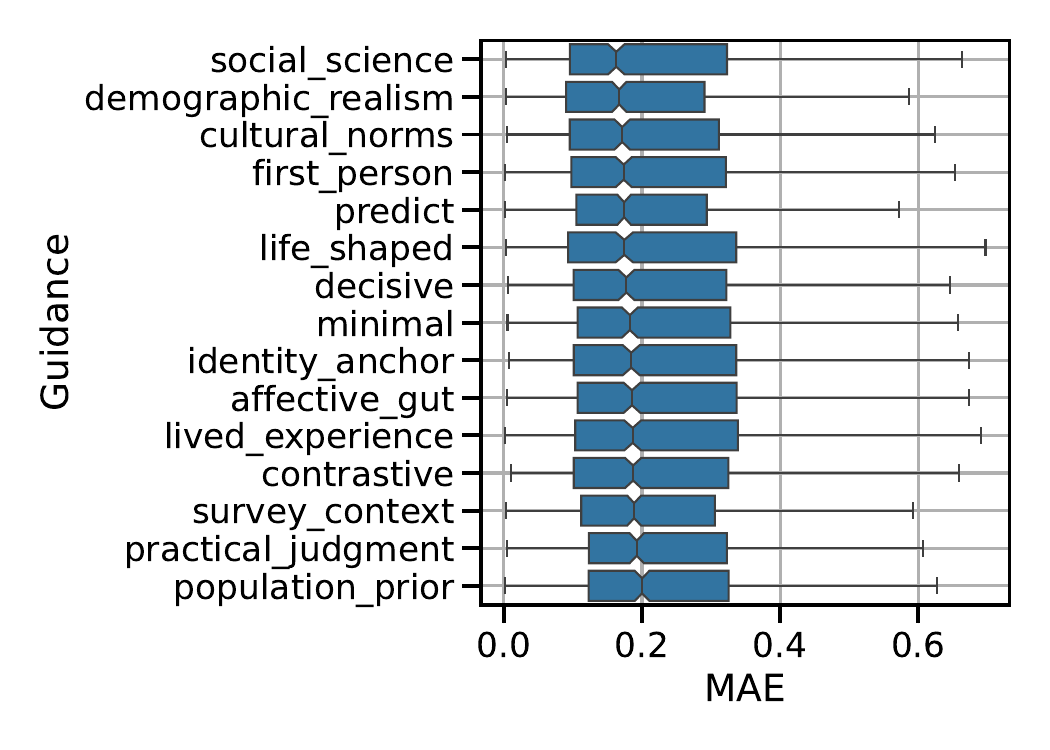}
\caption{\label{fig:guidance_comparison}Mean absolute error (MAE) between model predictions and human survey responses across countries, models, and prompting methods. Each boxplot shows the distribution of errors for one guidance variant.}
\end{figure}

Part of the persona prompt consists of a guidance instruction that asks the LLM to answer in accordance with the assigned persona. To assess how sensitive country prompting is to the wording of this instruction, we evaluate several guidance variants on the descriptor-question set (see Appendix \ref{sec:guidance_templates}). \autoref{fig:guidance_comparison} shows that guidance wording has only a modest effect on performance. Among the tested variants, \texttt{social\_science} performs best. This variant frames the task as a social-science simulation and instructs the model to answer as someone living in the specified country, adopting that person's values as its own.

\subsection{Descriptor Robustness}\label{sec:descriptor_robustness}

\begin{figure}
\centering
\includegraphics[width=.7\linewidth]{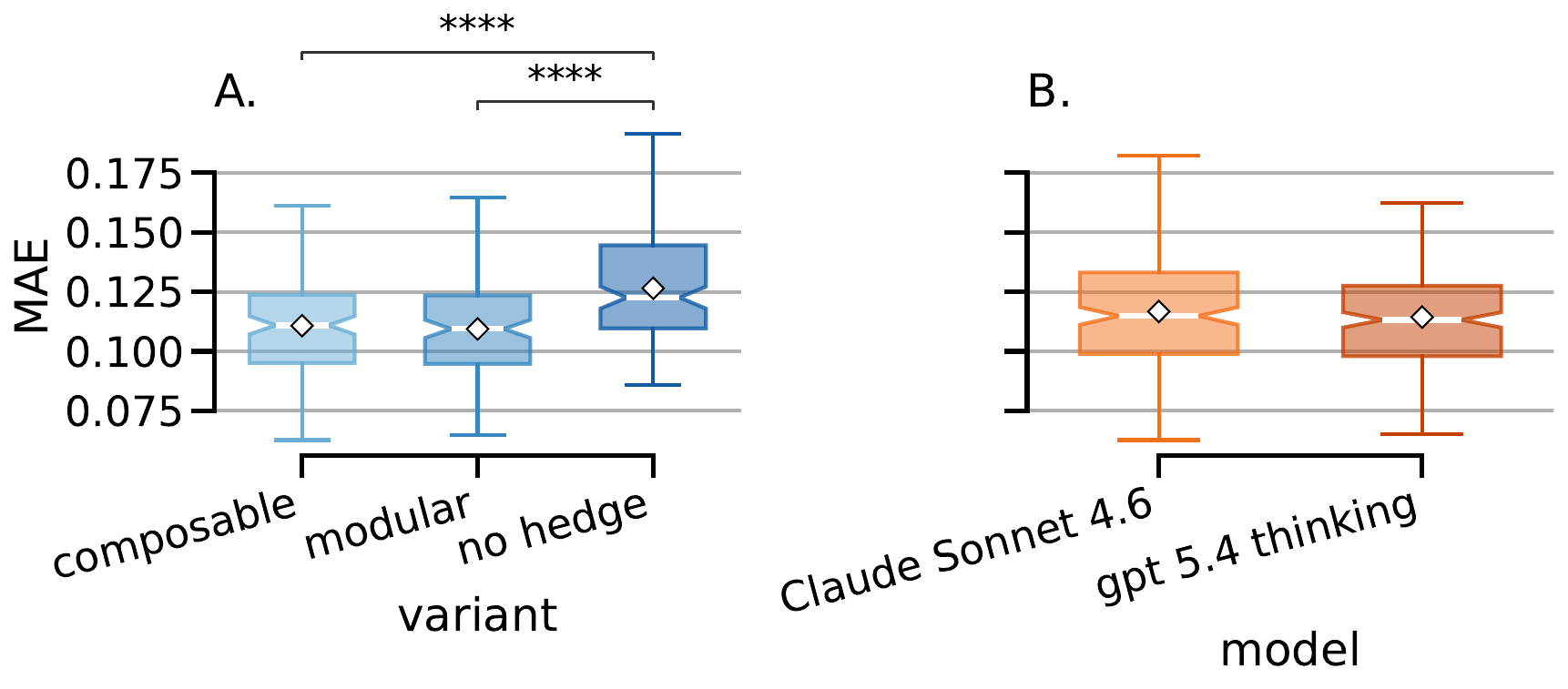}
\caption{\label{fig:error_distributions_generators}Mean absolute error (MAE) of value-based personas as a function of the generator choices. Significance annotations indicate Benjamini-Hochberg-corrected two-sided Mann-Whitney U tests; **** denotes $p<10^{-4}$.}
\end{figure}

Value-based personas are initialized with a sequence of natural language descriptors. We evaluate the impact of the generation choices in \autoref{fig:error_distributions_generators}. In particular, there are two main choices. First, which LLM is used to generate the descriptors. We evaluate here either GPT5.4 or Claude Sonnet 4.6, and show that there is no significant difference in the performance of value-based personas based on which LLM was used to generate the descriptors. 

Second, we vary the instruction provided to the descriptor-generating LLM. We consider three variants (see \autoref{tab:descriptor_prompts}) designed to test whether descriptor performance depends only on capturing the target response, or also on limiting unintended inferences. The \emph{no-hedge} prompt asks the generator to produce descriptors that encourage the target response, but does not explicitly restrict broader associations. We expect this variant to perform worse, because such descriptors may overspecify the persona by activating traits that are correlated with the response but not part of the sampled value profile.

The \emph{modular} and \emph{composable} prompts address this issue by instructing the generator to hedge against the spurious associations an LLM would most naturally draw from each response. For example, a descriptor corresponding to opposition to homosexuality should not necessarily imply a broader conservative identity, and a descriptor corresponding to having voted should not necessarily imply high political engagement. The composable prompt further emphasizes that descriptors should remain self-contained and coherent when combined across many survey items.

The results show that descriptor wording matters primarily through the use of hedging. The no-hedge condition performs worse than the modular and composable variants, indicating that descriptors which only encourage the target response can introduce unintended associations that degrade prediction accuracy. By contrast, the modular and composable prompts are not significantly different from one another. This suggests that explicitly hedging against spurious inferences is important, while the precise formulation of the hedged descriptor prompt is less critical.

\begin{longtable}{@{} l p{11cm} @{}}
\toprule
\textbf{Variant} & \textbf{Prompt} \\
\midrule
\endhead

Composable &
\small
For every response option, craft a one-sentence persona snippet starting with ``You.'' Each snippet should nudge an LLM toward giving that specific answer while suppressing broader inferences the model might otherwise make. The snippets must be self-contained and composable---mixing and matching one per question should produce a coherent, accurate persona.
\\[1.2em]

Modular &
\small
For each possible response, provide a short sentence describing the person who gave that answer. The descriptions should be modular, such that one descriptor per question can be combined into a coherent role-play persona.

Each descriptor should trigger the target response while hedging against broader associations the LLM might naturally infer. For example, opposition to homosexuality should not necessarily imply broader conservative views, and reporting that one has voted should not necessarily imply strong political engagement.

Start each descriptor with ``You.''
\\[1.2em]

No hedge &
\small
For each possible response, provide a short sentence describing the person who gave that answer. The descriptions should be composable, such that one descriptor per question can be combined into a coherent role-play persona.

Each descriptor should trigger the target response.

Start each descriptor with ``You.''
\\

\bottomrule
\caption{Descriptor-generation prompt variants used in the robustness analysis.}\label{tab:descriptor_prompts}
\end{longtable}

\subsection{Value Contributions Vary Across Countries}

\begin{figure*}
\centering
\includegraphics[width=1\linewidth]{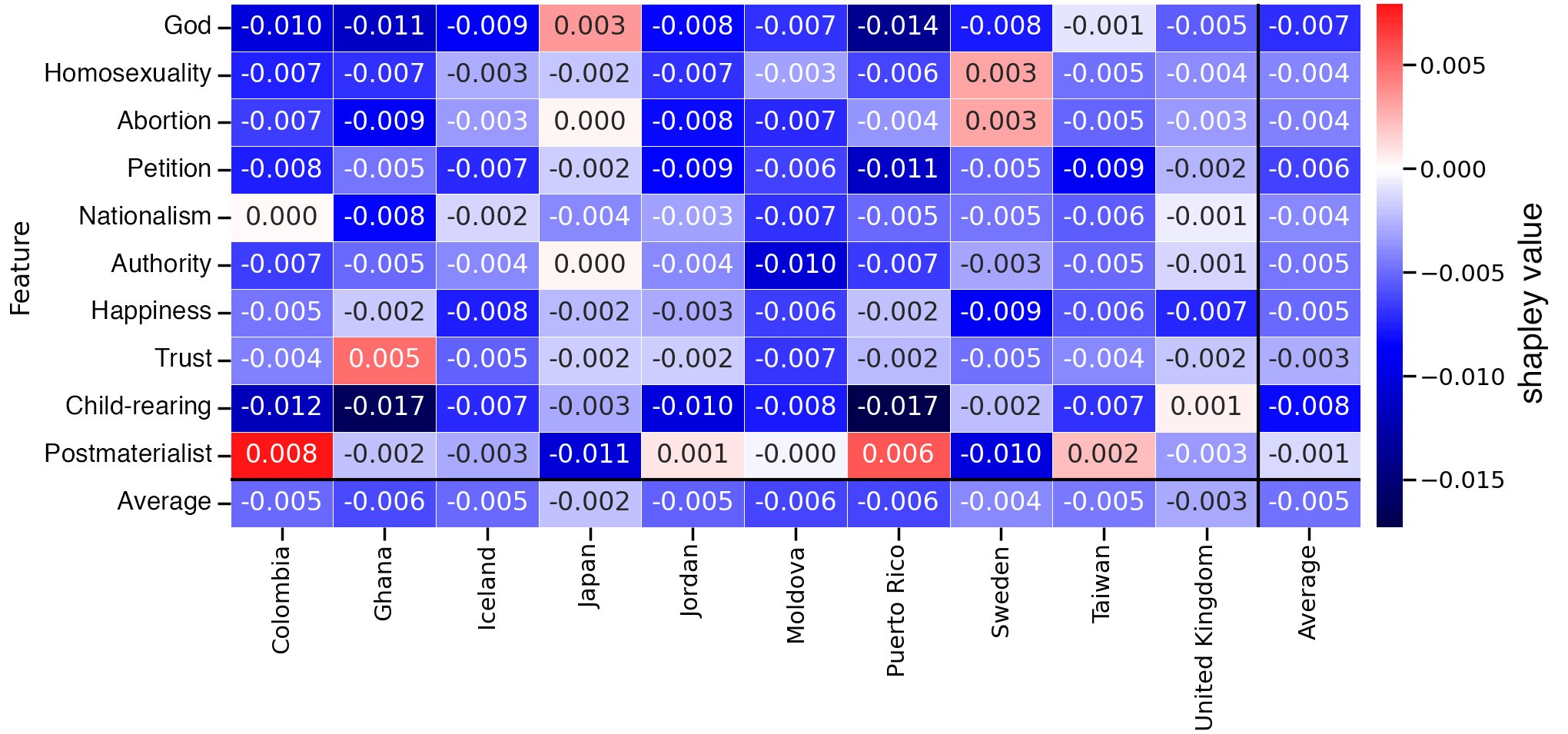}
\caption{\label{fig:shapley}Per-item Shapley values across countries. Negative values indicate that including the item improves performance (lower MAE), while positive values indicate a degradation.}
\end{figure*}

While our value-based personas are grounded in the same 10 items \cite{inglehart2005modernization} use to construct their cultural map, their relative contribution to predictive performance may differ across items and countries. We therefore quantify their impact using the Shapley values shown in \autoref{fig:shapley}.

First, contributions vary substantially across countries. The same item can improve performance in some contexts while degrading it in others. For instance, the postmaterialist index strongly improves performance for Japan and Sweden, but has the opposite effect for Colombia and Puerto Rico.

Second, despite this heterogeneity, all items contribute positively on average, indicating that each adds useful information to the model. 

Finally, aggregating across items, all countries benefit on average from their inclusion, with consistently negative mean contributions. This supports the hypothesis that the Inglehart-Welzel value framework provides a meaningful signal for steering LLM predictions across national contexts.

\subsection{Per-Question Robustness of Persona Prompting}

\begin{figure}
\centering
\includegraphics[width=.7\linewidth]{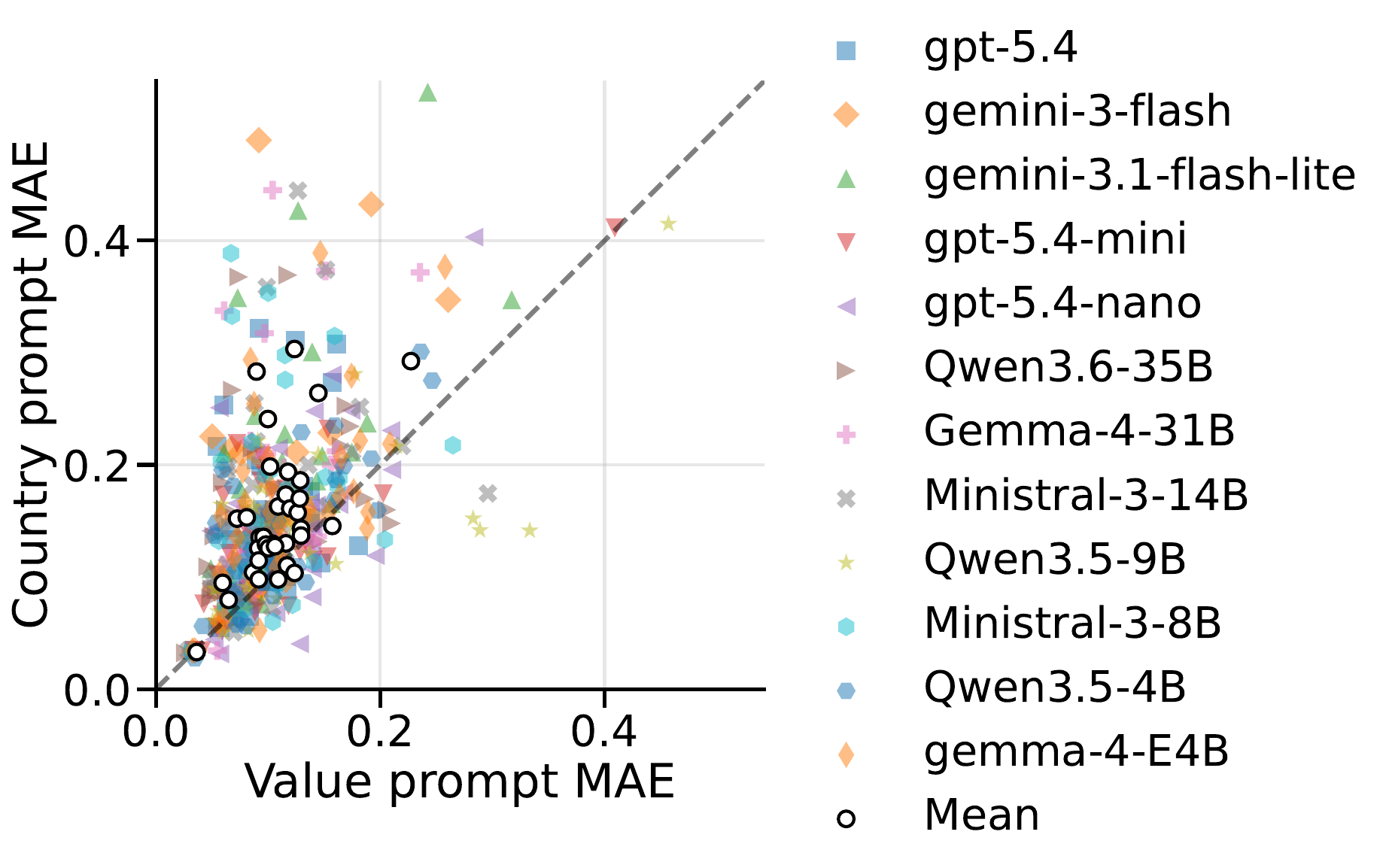}
\caption{\label{fig:error_comparison}Per-question comparison of MAE for value prompting versus country prompting. Each point corresponds to a model-question pair, with colored points indicating individual models, and black points showing the average across models. Points above the diagonal indicate higher error for country prompting. }
\end{figure}

To determine whether improvements are driven by a small subset of questions or are broadly distributed, we compare per-question errors between persona prompting and country prompting in \autoref{fig:error_comparison}.

We find that the majority of points lie above the diagonal, indicating that persona prompting outperforms country roleplay across a wide range of questions. This demonstrates that the observed improvements are systematic rather than driven by a small number of favorable cases.

\subsection{Impact of Temperature Scaling}\label{sec:temperature_all_curves}

\begin{figure*}[ht!]
\centering
\includegraphics[width=1\linewidth]{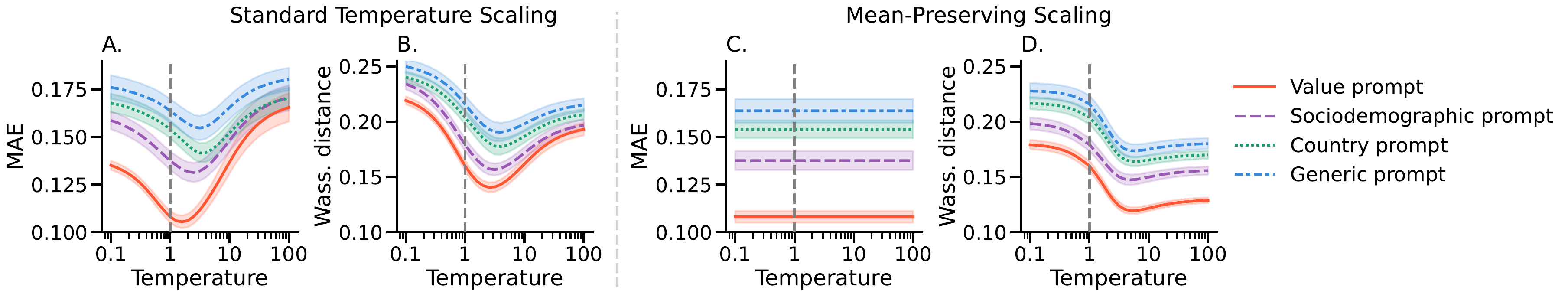}
\caption{\label{fig:temperature_all_curves}MAE and Wasserstein distance as a function of temperature for the two methods described in 
Section \ref{sec:tilting}. Panels A and B report the values for standard temperature scaling (as typically used when sampling from LLMs), while C and D give the MAE (constant by design) and Wasserstein distance for mean-preserving temperature scaling. The vertical dashed line indicates the default temperature of $1$. Shaded regions indicate 95\% confidence intervals. }
\end{figure*}

A common strategy for extracting survey responses from LLMs is to select the highest-probability response, corresponding to the zero-temperature limit. As discussed in \autoref{sec:tilting}, however, changing the temperature affects not only the dispersion of the response distribution but also its expected response. However, it is not clear a priori whether selecting the model's highest probability response is optimal.

To test this, \autoref{fig:temperature_all_curves}.A reports MAE as a function of temperature. We find that decreasing the temperature worsens performance across all prompting methods. Notably, the temperature of $0.3$ used by \cite{miranda2025simulating} performs worse than the model's default temperature. This suggests that making responses more deterministic can suppress useful uncertainty and degrade aggregate prediction accuracy.

\section{Supplementary figures}

\begin{figure*}
\centering
\includegraphics[width=1\linewidth]{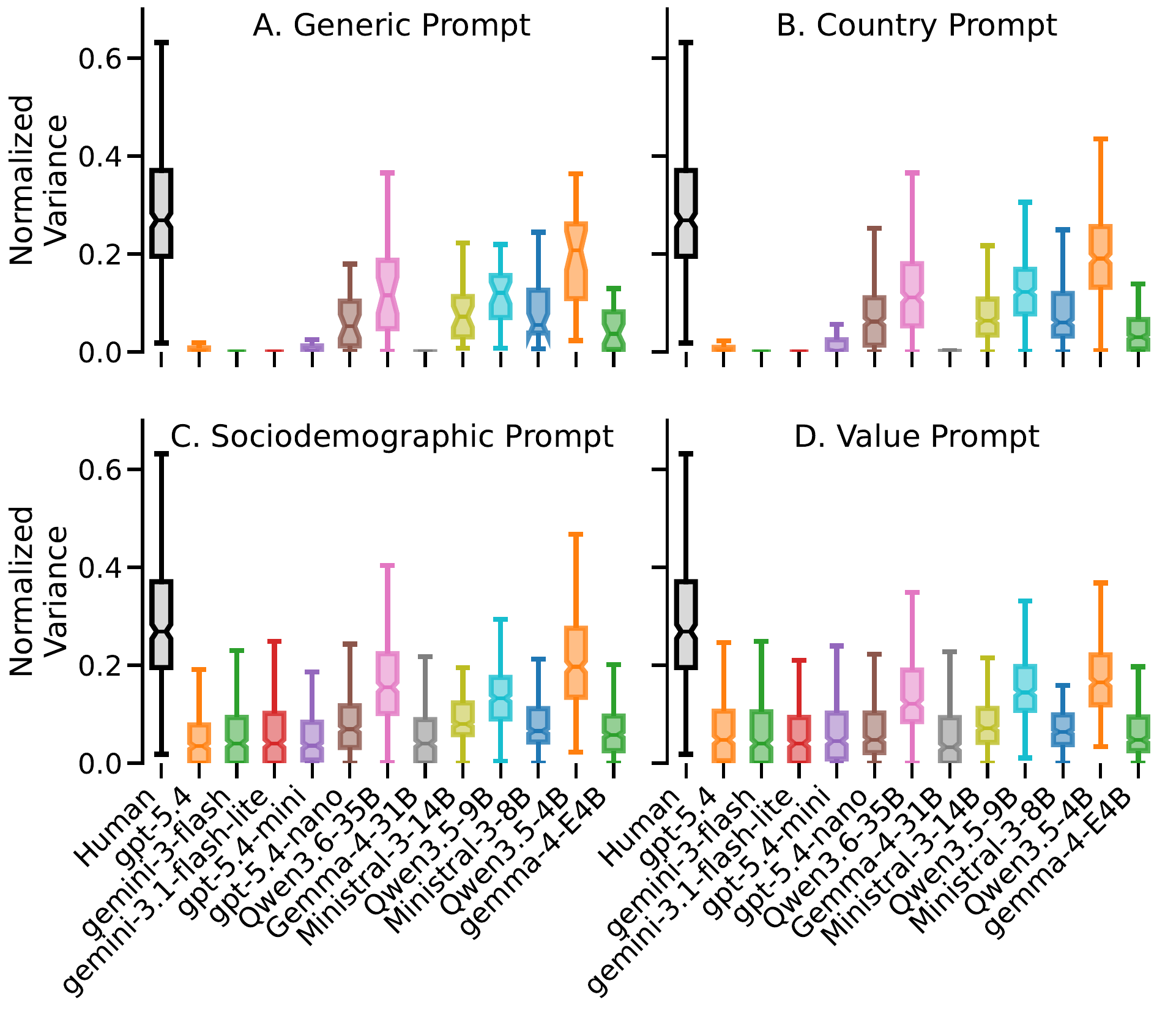}
\caption{\label{fig:full_peak_distributions}Distribution of normalized variance by method, compared to human responses. Each boxplot summarizes variance values across all question-country pairs. }
\end{figure*}

\begin{figure*}
\centering
\includegraphics[width=1\linewidth]{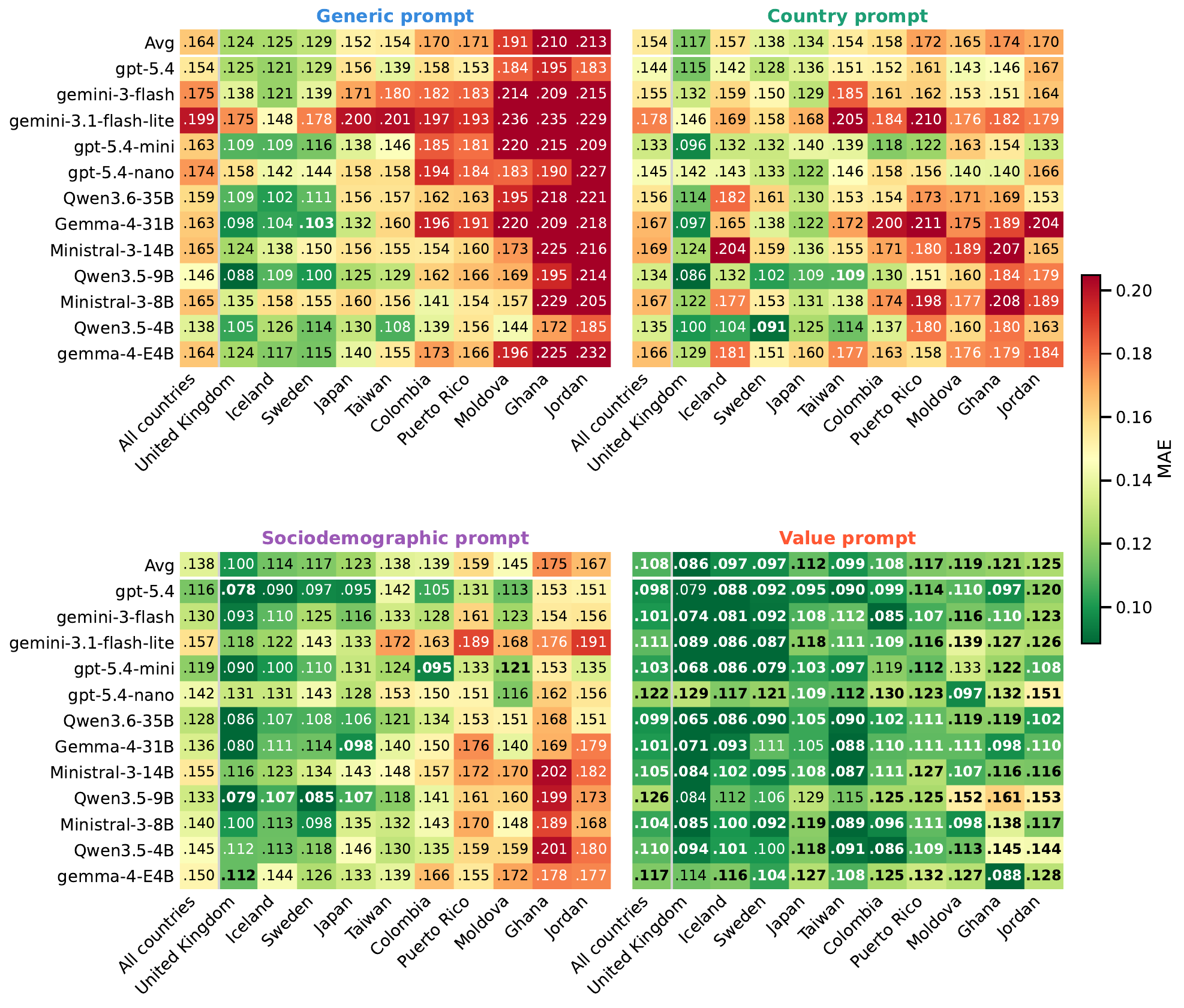}
\caption{\label{fig:heatmap_mae}Mean Absolute Error (MAE) of model predictions relative to human survey responses across countries and prompting methods.
Each panel corresponds to a prompting method (Generic Roleplay, Country Roleplay, and Persona Prompt), with rows representing individual language models and an ensemble average, and columns representing countries sorted in ascending order of Generic Roleplay MAE.
Cell values report the MAE averaged across all questions.
Bold values indicate the best-performing method for a given model-country pair; an underline further denotes that this advantage is statistically significant ($p < 0.05$, Wilcoxon signed-rank test across methods).
The top row reports the MAE across all models for each country. The first column reports the MAE across all countries for each model.}
\end{figure*}

\end{document}